\documentclass[letterpaper]{article} % DO NOT CHANGE THIS
\usepackage{aaai25}  % DO NOT CHANGE THIS
\usepackage{times}  % DO NOT CHANGE THIS
\usepackage{helvet}  % DO NOT CHANGE THIS
\usepackage{courier}  % DO NOT CHANGE THIS
\usepackage[hyphens]{url}  % DO NOT CHANGE THIS
\usepackage{graphicx} % DO NOT CHANGE THIS
\usepackage{natbib}  % DO NOT CHANGE THIS AND DO NOT ADD ANY OPTIONS TO IT
\usepackage{caption} % DO NOT CHANGE THIS AND DO NOT ADD ANY OPTIONS TO IT
\usepackage{newfloat}
\usepackage{listings}
\DeclareCaptionStyle{ruled}{labelfont=normalfont,labelsep=colon,strut=off} % DO NOT CHANGE THIS
\usepackage{booktabs}
\usepackage{amsmath}
\usepackage{amssymb}
\usepackage{bm}
\usepackage{dsfont} % For indicator function
\usepackage{algorithm2e}

\newcommand{\vctr}[1]{\bm{#1}}
\newcommand{\mtrx}[1]{\bm{#1}}

\DeclareMathOperator{\similarity}{sim}
\newcommand{\gate}{\textup{gate}}
\newcommand{\spanup}{\textup{span}}
\newcommand{\referup}{\textup{refer}}

\newcommand{\sizeup}{\textup{size}}

\newcommand{\dontcarevalue}{\texttt{dontcare}}
\newcommand{\informvalue}{\texttt{inform}}
\newcommand{\refervalue}{\texttt{refer}}

\title{Learning from Noisy Labels via Self-Taught On-the-Fly Meta Loss Rescaling}

\author {
    Michael Heck, %\textsuperscript{\rm 1},
    Christian Geishauser, %\textsuperscript{\rm 1},
    Nurul Lubis, %\textsuperscript{\rm 1},
    Carel van Niekerk, \\%\textsuperscript{\rm 1},
    Shutong Feng, %\textsuperscript{\rm 1},
    Hsien-Chin Lin, %\textsuperscript{\rm 1},
    Benjamin Matthias Ruppik, %\textsuperscript{\rm 1},
    Renato Vukovic, %\textsuperscript{\rm 1},
    Milica Ga\v{s}i\'{c}%\textsuperscript{\rm 1}
}
\affiliations {
    Heinrich Heine University Düsseldorf\\
    \{heckmi,geishaus,lubis,niekerk,fengs,linh,ruppik,revuk100,gasic\}@hhu.de
}

\begin{document}

\maketitle

\begin{abstract}
Correct labels are indispensable for training effective machine learning models. However, creating high-quality labels is expensive, and even professionally labeled data contains errors and ambiguities. Filtering and denoising can be applied to curate labeled data prior to training, at the cost of additional processing and loss of information. An alternative is on-the-fly sample reweighting during the training process to decrease the negative impact of incorrect or ambiguous labels, but this typically requires clean seed data. In this work we propose unsupervised on-the-fly meta loss rescaling to reweight training samples. Crucially, we rely only on features provided by the model being trained, to learn a rescaling function in real time without knowledge of the true clean data distribution. We achieve this via a novel meta learning setup that samples validation data for the meta update directly from the noisy training corpus by employing the rescaling function being trained.
Our proposed method consistently improves performance across various NLP tasks with minimal computational overhead. Further, we are among the first to attempt on-the-fly training data reweighting on the challenging task of dialogue modeling, where noisy and ambiguous labels are common. Our strategy is robust in the face of noisy and clean data, handles class imbalance, and prevents overfitting to noisy labels. Our self-taught loss rescaling improves as the model trains, showing the ability to keep learning from the model's own signals. As training progresses, the impact of correctly labeled data is scaled up, while the impact of wrongly labeled data is suppressed.
\end{abstract}

\begin{links}
     \link{Code}{https://gitlab.cs.uni-duesseldorf.de/general/dsml/storm-public}
     %\link{Datasets}{https://aaai.org/example/datasets}
     %\link{Extended version}{https://aaai.org/example/extended-version}
\end{links}

% ================================
\section{Introduction}
\label{sec:introduction}
% ================================

High-quality training data is paramount for developing accurate, safe, and reliable machine learning models. Real-world datasets, however, often contain noise, ambiguities, biases, and errors.  Ever larger models increase the need for automatic training data generation, exacerbating label noise issues. Inconsistencies in data degrade model training, limit performance and hinder generalizability~\cite{6685834}.
The complexity of human language renders natural language processing (NLP) labeling tasks inherently challenging, even for experienced annotators~\cite{marcus-etal-1993-building}. Inter-annotator agreement on labeling tasks depends on factors like task description and guidance, annotator skills and knowledge, and level of attention~\cite{10.1007/978-3-642-19400-9_14}.

One approach to mitigating noisy labels is sample selection, which involves identifying and removing noisy samples before training or updating the model~\cite{shen2019learningbadtrainingdata,chen2019understandingutilizingdeepneural,9008796}.
This method exploits the overfitting tendency of DNNs to detect noisy samples through model weight trajectories or loss cutoffs~\cite{yao2020searchingexploitmemorizationeffect,li2019gradientdescentearlystopping,shen2019learningbadtrainingdata}. Advanced approaches use collaborative multi-network learning~\cite{han2018coteachingrobusttrainingdeep,jiang2018mentornetlearningdatadrivencurriculum}. Despite its effectiveness, sample selection requires significant computational resources, struggles with over-filtering, and
makes permanent filtering errors early in the process, causing information loss.

Another tactic is loss adjustment, which modifies sample losses during training, allowing to fully explore the noisy dataset. This can be achieved by learning noise transition matrices~\cite{patrini2017makingdeepneuralnetworks}, correcting labels~\cite{reed2015trainingdeepneuralnetworks}, estimating reweighting functions~\cite{7929355} or via meta learning~\cite{finn2017modelagnosticmetalearningfastadaptation}. Loss \emph{rescaling} adjusts sample losses to control their effect on the model training on-the-fly. Examples of methods for DNNs are importance reweighting and active bias~\cite{Liu_2016,chang2018activebiastrainingaccurate}. Other approaches utilize additional models to identify noisy samples during training~\cite{9577967}. Despite their versatility, these methods typically require clean data to estimate ``true'' distribution of uncorrupted data, and rely on predefined adjustment functions and hyperparameters related to the expected noise type.

Meta learning offers a solution by automating loss rescaling, thus eliminating manual function definition. This higher-order learning estimates noise type agnostic rescaling functions using meta models or unified neural models~\cite{dehghani2017learninglearnweaksupervision,shu2019metaweightnetlearningexplicitmapping,li2019learninglearnnoisylabeled,ren2019learningreweightexamplesrobust}. Yet, they still depend on clean validation data to learn the rescaling function or the meta objective.

\emph{But what if clean data is not available?} Oftentimes clean data simply does not exist and its procurement is time consuming, expensive and itself prone to errors.

Adaptive gradient-based outlier removal (AGRA)~\cite{Sedova_2023} sidesteps the need for clean validation data by using gradient comparisons to decide on-the-fly whether a sample is useful in the current stage of training. However, its binary decisions, reliance on gradient similarity and lack of meta learning limit AGRA's flexibility and robustness.

In this paper, we introduce \textbf{STORM} (\textbf{S}elf-\textbf{T}aught \textbf{O}n-the-fly \textbf{R}escaling via \textbf{M}eta loss), a flexible loss rescaling method for learning from noisy labels.
Our contributions are as follows:

\begin{description}
    \item[A novel meta learning scheme] called \emph{meta loss rescaling} that eliminates the need for clean validation data by rescaling both the loss in the inner loop and the meta loss in the outer loop, using \emph{noisy} validation data.
    \item[A \emph{flexible} loss rescaling] that dynamically decides how much importance to assign to a sample at each training stage and keeps learning from the model's own signals.
    \item[An \emph{efficient} loss rescaling] that uses features based on sample losses and prediction probabilities instead of sample gradients, reducing computational complexity.
    \item[A \emph{robust} loss rescaling] that handles class imbalance, different types of noise, and prevents overfitting.
    \item[An application to dialogue modeling] as an underexplored use case for loss rescaling, markedly improving performance in this noise sensitive task.
    \item[An extensive empirical evaluation] on various NLP tasks that validates STORM's ability to identify noisy and ambiguous samples with high recall and low false positives.
\end{description}

% ================================
\section{Self-Taught On-the-Fly Meta Loss Rescaling}
\label{sec:method}
% ================================

\begin{figure}[t]
	\centering
	\includegraphics[width=1.0\linewidth]{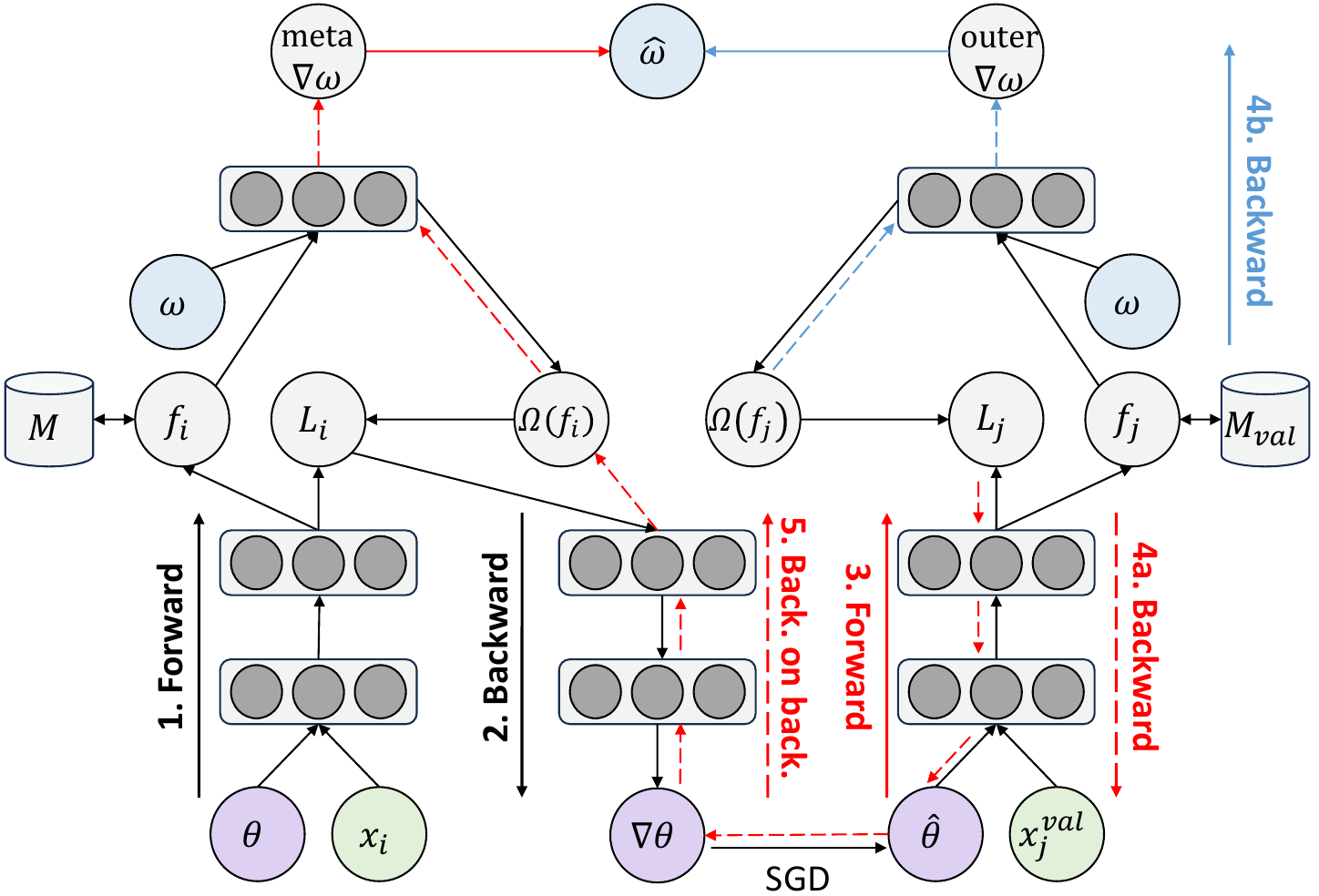}
	\caption{Schematic of STORM. Numbered black arrows correspond to the inner loop updates, i.e., learning $\Theta$. Numbered red and blue arrows correspond to the outer loop updates, i.e., meta learning $\Omega$. Dashed arrows indicate gradient flows.}
	\label{fig:storm}
\end{figure}

STORM aims to decrease the impact of noisy labels and increase the impact of correctly labeled samples, without prior assumptions about noise distribution or using clean reference data. Rescaling is done on-the-fly, starting with a freshly initialized model without prior knowledge of the target task. Using meta loss rescaling to learn from noisy labels avoids information loss, as it does not omit data from training entirely and decisions are not static. At each training epoch, upon revisiting training data, the rescaling is adjusted according to the current state of the model being trained such that the negative impact of label noise is minimized. This results in (1) the model benefiting differently from the same sample at each epoch, and (2) the rescaling function continuously updating via self-teaching based on the model's grasp of the data.
Fig.~\ref{fig:storm} is an illustration of STORM, and Alg.~\ref{alg:storm} outlines the algorithm.

\subsection{Optimization Problem}

We use classification as a representative machine learning task. We assume to only have access to a training set $\mathfrak{X} = \{(x_i,\tilde{y}_i)\}_{i=1}^N$ with \emph{noisy} labels $\tilde{y}_i$. Let $\Theta_\theta(\cdot): \mathfrak{X} \rightarrow \mathbb{R}^C$ be a model we train to solve a $C$ way classification task.
The parameters $\theta$ are subject to optimization via minimizing a loss function $\ell_i = \mathcal{L}_\theta(\hat{y}_i, \tilde{y}_i)$ for each pair of model prediction $\hat{y}_i = \text{argmax}_c \Theta_\theta(x_i)$ and noisy label $\tilde{y}_i$ for input $x_i$.

In regular training, we minimize $\sum^N_{i=1} \mathcal{L}_\theta(\hat{y}_i, \tilde{y}_i)$. With STORM, we learn a rescaling function $\Omega_\omega(\cdot)$ such that samples are effectively reweighted, i.e., we optimize $\theta$ by minimizing a \emph{weighted} loss,
\begin{equation}
    \theta^* = \text{argmin}_\theta \sum^M_{i=1} \Omega_\omega(\vctr{f}_i) \cdot \mathcal{L}_\theta(\hat{y}_i, \tilde{y}_i).
    \label{eq:wloss}
\end{equation}
Our rescaling function expects input features $\vctr{f}_i$ that are computed directly from signals of the model being trained by conducting forward passes through $\Theta$. The optimal $\omega$ is not known and needs to be learned by minimizing the \emph{meta} loss, i.e., 
\begin{equation}
    \omega^* = \text{argmin}_\omega \sum^M_{j=1} \mathcal{L}_{\theta^*}(\hat{y}_{j}, \tilde{y}_{\text{val},j}),
\end{equation}
where $\mathfrak{V} = \{(x_{\text{val},j}, \tilde{y}_{\text{val},j})\}_{j=1}^M$ are samples from a validation set. For STORM, it is $\mathfrak{V} = \mathfrak{X}$, therefore $\tilde{y}_{\text{val},j}$ is a \emph{noisy} label. It becomes apparent that optimizing $\Theta$ and $\Omega$ is a chicken-and-egg problem. Meta learning allows us to continuously update $\Omega$ on-the-fly, i.e., during the training of $\Theta$, in a single optimization loop. 

\subsection{Learning the Model $\Theta$}

Without loss of generalization, we define
\begin{equation}
    \Theta_\theta(x_i) = \text{softmax}(\text{Classify}_{\theta_\text{c}}(\text{Enc}_{\theta_\text{e}}(x_i))) \in \mathbb{R}^C,
    \label{eq:theta}
\end{equation}
$\text{Enc}(\cdot)$ is an input encoder such as a transformer network, and $\text{Classify}(\cdot)$ is a trainable model such as a single linear layer, a multi layer perceptron or a deep neural network. 
It is $(\theta_\text{c}, \theta_\text{e}) \in \theta$.

Meta learning -- also referred to as ``learning to learn'' -- involves two levels of learning: an inner loop and an outer loop. In the inner loop, the model $\Theta$ is trained on a specific task. The outer loop trains meta parameters by evaluating $\Theta$’s performance on a meta validation set and backpropagating through a rolled-out inner loop's gradient graph.
In the inner loop of meta learning, we update $\theta$ via stochastic gradient descent (SGD) using the weighted loss of Equation~\ref{eq:wloss}, which helps the model to focus on more relevant samples within a mini-batch $\mathfrak{B} \in \mathfrak{X}$. $\omega$ is initialized such that $\Omega_\omega(\mathfrak{B})$ produces uniformly distributed weights across samples in the mini-batch, i.e., initially, the inner loop approximates regular training without rescaling. 

\subsection{Input Features to $\Omega$}
\label{sec:method:ssec:feats}

The input features $\vctr{f}_i$ to $\Omega$ for a single sample $x_i$ are obtained from $\Theta$. We can perform a forward pass multiple times with random dropout to generate variance in the model output. The probability and loss of forward pass $g$ out of $G$ passes is denoted as $p_{i,g}$ and $\ell_{i,g}$, respectively. We store these two fundamental features in a LIFO memory $\mathfrak{M}_\mathfrak{D}$ for all samples in a batch $\mathfrak{B} = \{(x_i,\tilde{y}_i)\}_{i=1}^B \in \mathfrak{D}$. The size of this memory is set by a parameter $m_{\sizeup}$. $\mathfrak{M}_\mathfrak{D}$ is continuously updated throughout training. With these two fundamental feature types, we compute additional features according to equations in Tab.~\ref{tab:feats}. We compute means and standard deviations for individual samples and across sample groups, which helps capture central tendencies and variability as measures for sample typicality. Kullback-Leibler (KL) divergence and overlap coefficient (OVL)~\cite{doi:10.1080/03610928908830127} are more sophisticated measures of distributional differences and similarities that help detect outliers. KL and OVL provide complementary information. KL measures divergence, is asymmetric and more sensitive to outliers, while OVL measures agreement and is symmetric. Lastly, a binary indicator (CAT) for model prediction and label agreement adds the factor of model performance to $\Omega$'s input, which helps identify challenging samples.
Our feature set is
\begin{equation}
\begin{aligned}
    \vctr{f}_i = (& \bar{\ell}_i, \bar{\ell}, \bar{\breve{\ell}}, \breve{\ell}_i, \breve{\ell}, \breve{\breve{\ell}},
             \bar{p}_i, \bar{p}, \bar{\breve{p}}, \breve{p}_i, \breve{p}, \breve{\breve{p}}, \\
           & \text{KL}_i, \bar{\text{KL}}, \bar{\breve{\text{KL}}},
             \text{OVL}_i, \bar{\text{OVL}}, \bar{\breve{\text{OVL}}}, \text{CAT}_i ).
\end{aligned}
\end{equation}
In practice, we separate the feature computation by target classes $C$, i.e., $\vctr{f}_i(\tilde{y}_i)$ is target class dependent. For readability, we use the shorthand $\vctr{f}_i$ for the remainder of the paper.

\RestyleAlgo{ruled}
\SetKwComment{Comment}{// }{}
\begin{algorithm}[t]
\caption{STORM}\label{alg:storm} 
\KwData{Initial $\theta$, initial $\omega$, noisy training batches $\mathfrak{X}$, noisy validation batches $\mathfrak{V}$, forward passes $G$} 
\KwResult{Trained model $\Theta_{\theta^*}$ and rescaler $\Omega_{\omega^*}$}
 \While{training continues}{
  \Comment{Inner loop learns $\Theta$}
  \For{each inner loop traversal}{
   $\mathfrak{B}$ $\gets$ SampleBatch($\mathfrak{X}$)\;
   $\mtrx{F}$ $\gets$ GetRescalerFeatures($\Theta_\theta$, $\mathfrak{B}$, $G$)\;
   $\ell$ $\gets$ Forward($\Theta_\theta$, $\mathfrak{B}$)\;
   $\nabla \theta$ $\gets$ Backward($\Omega_\omega(\mtrx{F})$, $\ell$)\;
   $\theta^*$ $\gets$ Optimize($\theta$, $\nabla \theta$)\;
   $\theta$ $\gets$ $\theta^*$\;
  }
  \Comment{Outer loop meta learns $\Omega$}
  $\mathfrak{B}_\text{val}$ $\gets$ SampleBatch($\mathfrak{V}$)\;
  $\mtrx{F}_\text{val}$ $\gets$ GetRescalerFeatures($\Theta_{\theta^*}$, $\mathfrak{B}_\text{val}$, $G$)\;
  $\ell_{\theta^*}$ $\gets$ Forward($\Theta_{\theta^*}$, $\mathfrak{B}_\text{val}$)\; % TODO: update ells in text?
  $\nabla_\text{meta} \omega$, $\nabla_\text{outer} \omega$ $\gets$ Backward($\Omega_\omega(\mtrx{F}_\text{val})$, $\ell_{\theta^*}$)\;
  $\omega^*$ $\gets$ Optimize($\omega$, $\nabla_\text{meta} \omega$, $\nabla_\text{outer} \omega$)\;
  $\omega$ $\gets$ $\omega^*$\;
 }
\end{algorithm}

\begin{table}[t]
    \small
    \begin{tabular}{ll}
        $\bar{o}_i = \texttt{mean} \left( \left\{ o_{i,g} \right\}_{g=1}^G \right)$ &
        $\breve{o}_i = \texttt{std} \left( \left\{ o_{i,g} \right\}_{g=1}^G \right)$ \\
        $\bar{o} = \texttt{mean} \left( \left\{ \bar{o}_b \right\}_{b=1}^B \right)$ &
        $\breve{o} = \texttt{std} \left( \left\{ \bar{o}_b \right\}_{b=1}^B \right)$ \\
        $\bar{\breve{o}} = \texttt{mean} \left( \left\{ \breve{o}_b \right\}_{b=1}^B \right)$ &
        $\breve{\breve{o}} = \texttt{std} \left( \left\{ \breve{o}_b \right\}_{b=1}^B \right)$ \\
    \end{tabular}
    \begin{tabular}{ll}
                    $\text{KL}_i = \mathcal{D}_\text{KL} \left( 
                        \mathcal{N} \left( \bar{\ell}_i, \breve{\ell}_i \right) \,\bigg\|\, 
                        \mathcal{N} \left( \bar{\ell}, \bar{\breve{\ell}} \right) 
                    \right)$ & 
        $\text{CAT}_i = \mathds{1}_{\hat{y}_i = \tilde{y}_i}$ \\
    \end{tabular}
    \begin{tabular}{l}
                    $\text{OVL}_i = \texttt{overlap} \left( 
                        \mathcal{N} \left( \bar{\ell}_i, \breve{\ell}_i \right), \,
                        \mathcal{N} \left( \bar{\ell}, \bar{\breve{\ell}} \right) 
                    \right)$ \\
    \end{tabular}
    \caption{Equations for individual and group-level feature computation based on losses $\ell_i$ and probabilities $p_i$.}
    \label{tab:feats}
\end{table}

\subsection{Meta Learning the Rescaling Function $\Omega$}
\label{sec:method:rescaling}

In the outer loop, we calculate the total \emph{meta} loss of a batch $\mathfrak{B}_\text{val}$ of samples from a validation set w.r.t. to the loss weights $\Omega_\omega(\mathfrak{B)}$ of the current training batch $\mathfrak{B}$ and compute its meta-gradients. For this, we roll out the inner loop's gradient graph to get the gradients $\nabla \theta$. Then we compute a backward-on-backward derivative given the meta loss to get second-order gradients for a full backward pass through the unrolled graph resulting in $\nabla_\text{meta}\omega$.

With this backward pass, we could update $\Omega$ w.r.t. the meta loss. Since we lack clean validation data for the meta update, we modify this step as follows: we sample $\mathfrak{B}_\text{val}$ from the \emph{noisy} training set and use $\Omega$ to rescale its individual sample losses in the outer loop, i.e.,
\begin{equation}
    \omega^* = \text{argmin}_\omega \sum^M_{j=1} \Omega_\omega(\vctr{f}_j) \cdot \mathcal{L}_{\theta^*}(\hat{y}_{j}, \tilde{y}_{\text{val},j}),
\end{equation}
also backpropagating through $\Omega$ with regular first-order derivatives resulting in $\nabla_\text{outer}\omega$. Therefore, $\omega$ is updated jointly via gradient accumulation of $\nabla_\text{meta}\omega$ and $\nabla_\text{outer}\omega$. In the meta-update step, only the rescaling function $\Omega$ undergoes parameter updates.
The rescaling function is implemented as a trainable neural network. Without loss of generalization, we design the rescaling function as
\begin{equation}
\begin{split}
    &[w_{0,i}, w_{1,i}] = \Omega_\omega(\vctr{f}_i) = \\
    &\quad\text{softmax}(\text{BN}_2(\text{L}_2(\text{ReLU}(\text{BN}_1(\text{L}_1(\vctr{f}_i))))) \in \mathbb{R}^2
\end{split}
\end{equation}
where $\text{L}_{(\cdot)}$ is a linear layer, $\text{BN}_{(\cdot)}$ is a batch normalization layer~\cite{ioffe2015batchnormalizationacceleratingdeep}, and $\text{ReLU}(\cdot)$ is the rectified linear unit activation function.
While the general design of this network is flexible, the last batch normalization layer is critical. Our rescaler has two target classes. We use the prediction $w_{0,i}$ as the loss rescaling weight and discard $w_{1,i}$. The batch normalization before the softmax layer ensures that the degenerate case is prevented where $w_{0,i} \approx 0 \text{ }\forall i$.
In practice, we learn a separate rescaling function for each target class of the model's training corpus $\mathfrak{X}$, i.e., $\Omega_{\omega,c}(\cdot) \forall c \in C$. This accounts for class imbalance and different feature distributions across classes. 

\subsection{Meta Learning from Noisy Data}

To increase robustness towards noisy training data in a meta learning setting, one uses a clean validation set for the meta update. The difference in distributions between training and meta validation set allows the model to learn a reweighting of data that is closer to the validation data distribution, thereby converging to a model better suited for such data.
In contrast, STORM operates without the need for access to clean data. Instead, validation batches are randomly sampled directly from the noisy training data at each update step. To establish a difference in data distributions, we rescale the validation loss using the same $\Omega$ that is currently being trained. 

The intuition is as follows. The inner and outer loop updates pursue complementary goals, creating a feedback loop that refines and improves weight predictions and model performance. The outer loop minimizes the total validation loss by rescaling the validation sample loss weights, which is most efficient if the loss of outlier samples is downscaled. Concurrently, the outer loop updates the weights used in the inner loop such that the model update is becoming more beneficial w.r.t performance on the validation set. Successful joint training of the model and minimization of the meta loss depend on increasing the weights of beneficial samples while decreasing those of less useful ones. Batch normalization plays a critical role in this process, as it prevents the predicted weights in the outer loop from collapsing to zero.
As the model utilizes more informative samples in the inner loop, weighting in the outer loop becomes more accurate. Conversely, as the outer loop downscales outlier samples, the model training increasingly focuses to minimize the validation loss of non-outlier samples. 

\subsection{Differences to AGRA}

AGRA is closely related to our work. It dynamically evaluates the utility of each training sample, making binary decisions at every training step based on the sample's prediction error. AGRA's decisions regarding individual samples may change as training progresses and are solely based on the gradients of the model being trained.
AGRA assumes noisy data $\mathfrak{X}$. It defines a comparison loss function $\tilde{\mathcal{L}}(x, y)$ for computing comparison gradients. For each $\mathfrak{B}$, a comparison batch $\mathfrak{B}_\text{comp}$ is sampled from $\mathfrak{X}$.
The sample gradients' similarity to the aggregated comparison gradient is:
\begin{equation}
    \similarity(i) = \frac{\nabla \tilde{\mathcal{L}}(x_i,y_i) \cdot \nabla \tilde{\mathcal{L}}(\mathfrak{B}_\text{comp})}{||\nabla \tilde{\mathcal{L}}(x_i,y_i)||_2 \cdot ||\nabla \tilde{\mathcal{L}}(\mathfrak{B}_\text{comp})||_2}.
\end{equation}
The assumption is: if gradients point in opposing directions, the sample may be harmful to model training given it's current state. $\theta$ is updated w.r.t. $\mathfrak{B}' = \{(x_i, y_i) \mid \similarity(i) \leq 0\}$.

Differences between our approaches include AGRA being gradient-based, requiring forward and backward passes for outlier removal, and applying a similarity measure for binary decisions. AGRA does not use meta learning, requires comparison batch sampling, and uses a class balancing heuristic.
STORM avoids extra backward passes, uses a learnable, meta-trained, data-driven rescaling function with continuous predictions. It does not require sampling a comparison batch, and uses a memory for balanced class-dependent statistics instead.
The similarities make both methods directly comparable, with distinctions likely affecting filtering and final model performance.

\subsection{Computational Complexity}
\label{sec:method:ssec:complexity}

STORM adds minimal computational overhead compared to regular meta learning. It involves two forward (Fig.~\ref{fig:storm}, steps 1 \& 3) and two backward (steps 2 \& 4b) passes, plus a backward-on-backward (step 4a) pass, with feature computations during steps 1 and 3 being negligible. Overall, meta learning has about triple the computational needs as regular training. Depending on $G$, steps 1 and 3 perform further computationally efficient forward passes through $\Theta$ without building a gradient graph.
The rescaling function $\Omega$ s a small multi-layer perceptron requiring negligible extra memory and computational resources.
Meta learning significantly increases memory consumption due to gradient graph unrolling, roughly doubling memory usage with a single inner loop traversal. Maintaining the memories $\mathfrak{M}_{(\cdot)}$ for feature computation is negligible.

% ================================
\section{Use Case: Dialogue State Tracking}
\label{sec:dst}
% ================================

A dialogue is a sequence of turns $\{(U_i, M_i)\}_{i=1}^T$, where $U_i$ is a user utterance and $M_i$ a preceding system utterance in natural language. Concepts of relevance for tracking are typically described in an ontology on the levels of domain (e.g., restaurant), slot (e.g., name) and value (e.g., Rosa's).  
In dialogue modeling, dialogue state tracking (DST) is the task of predicting user's intent from natural language input and keeping track of user's goal throughout the conversation as part of a dialogue state~\cite{YOUNG2010150}. That is, DST predicts at each turn the presence of domain-slot pairs $\{S_i\}_{i=1}^S$, their values, and updates the dialogue state. Accurate DST is crucial for high-performing dialogue systems. While specialized systems outperform LLM-based solutions in knowledge and privacy-intensive tasks, they require data with fine-grained and consistent labels~\citep{shen2024protosimi}.

Dialogue datasets can be grouped into machine-machine, machine-human and human-human categories. The latter is preferable for modeling human behavior but comes with the highest labeling costs and requirements. The Wizard-of-Oz (WOZ) framework~\cite{10.1145/357417.357420} generates and labels natural human-human interactions and has produced a range of widely used datasets~\cite{wen-etal-2017-network,budzianowski-etal-2018-multiwoz} for developing task-oriented dialogue systems~\cite{balaraman-etal-2021-recent}.

Despite rigorous selection of annotators, dialogue annotation presents significant challenges, especially w.r.t. consistency on intra- and inter annotator level~\cite{stolcke-etal-2000-dialogue,10.1093/jos/17.1.7}. According to~\citet{budzianowski-etal-2018-multiwoz}, annotating dialogue acts, which consist of an intent and a domain-slot-value triplet, is the most challenging part of dialogue data collection. Errors in dialogue act annotations typically lead to errors in dialogue state annotations, both crucial for dialogue system development.
Estimates for erroneous labels in MultiWOZ 2.1~\cite{budzianowski-etal-2018-multiwoz} -- one of the most widely used dialogue benchmarks -- range from 17\% on dialogue state, 22\% on slot, 31\%-41\% on turn and 28\%-65\% on dialogue level. Despite rigorous labeling processes, considerable noise remains in the dataset.

\begin{table*}[t]
  \centering
  \small
  \begin{tabular}{lccccccc}
    \toprule
     & Youtube & SMS  & CoLA & MRPC & RTE & MultiWOZ & Avg.\\
     & (Acc.) & (F1) & (Matth.) & (F1) & (F1) & (JGA) & - \\
    \midrule\midrule
    \textit{Clean labels} \\
    \midrule
    No rescaling                 & 93.4$\pm$0.3 & 93.7$\pm$1.3 & 62.0$\pm$1.8 & 91.7$\pm$1.1 & 78.9$\pm$1.9 & / & 83.9 \\
    AGRA                         & 94.0$\pm$0.3$^{\Uparrow}$ & 93.6$\pm$0.6 & 58.2$\pm$2.1$^{\Downarrow}$ & 90.7$\pm$0.8$^{\Downarrow}$ & 74.8$\pm$1.7$^{\Downarrow}$ & / & 82.3 \\
    \midrule
    STORM (ours)                 & 93.8$\pm$0.7 & 92.1$\pm$2.3 & 61.6$\pm$1.2 & 91.2$\pm$0.9$^{\downarrow}$ & 77.3$\pm$2.2$^{\downarrow}$ & / & 83.2 \\
    \midrule\midrule
    \textit{Noisy labels} & \multicolumn{5}{c}{\textit{(uniform noise 10\%)}} & \\
    \midrule
    No rescaling                 & 92.9$\pm$0.7 & 76.9$\pm$0.5 & 57.3$\pm$1.7 & 89.7$\pm$1.0 & 73.0$\pm$2.2 & / & 78.0 \\
    \midrule
    STORM (ours)                 & 93.5$\pm$0.4$^{\uparrow}$ & 88.1$\pm$1.0$^{\Uparrow}$ & 59.3$\pm$1.7$^{\Uparrow}$ & 90.5$\pm$1.0$^{\Uparrow}$ & 73.7$\pm$2.0 & / & 81.0 \\
    \midrule\midrule
    \textit{Noisy labels} & \multicolumn{5}{c}{\textit{(uniform noise 20\%)}} & \\
    \midrule
    No rescaling                 & 92.9$\pm$1.1 & 71.6$\pm$3.2 & 52.6$\pm$2.6 & 87.4$\pm$1.7 & 71.4$\pm$2.5 & / & 75.2 \\
    \midrule
    STORM (ours)                 & 93.6$\pm$0.9 & 85.3$\pm$1.3$^{\Uparrow}$ & 55.4$\pm$1.3$^{\Uparrow}$ & 89.3$\pm$1.2$^{\Uparrow}$ & 72.6$\pm$2.0 & / & 79.2 \\
    \midrule\midrule
    \textit{Noisy labels} & \multicolumn{5}{c}{\textit{(uniform noise 30\%)}} & \textit{(real noise)} \\
    \midrule
    No rescaling                 & 89.5$\pm$1.5 & 63.6$\pm$4.5 & 49.8$\pm$2.5 & 84.6$\pm$2.1 & 67.9$\pm$1.6 & 65.0$\pm$0.8 & 70.1 \\
    AGRA                         & 89.7$\pm$2.0 & 70.0$\pm$4.5$^{\Uparrow}$ & 47.7$\pm$3.2 & 85.9$\pm$2.9 & 68.9$\pm$0.5$^{\uparrow}$ & 62.1$\pm$0.6$^{\Downarrow}$ & 70.7 \\
    Meta learning w/ clean val.  & 89.1$\pm$2.1 & 74.9$\pm$7.1$^{\Uparrow}$ & 49.6$\pm$3.1 & 86.2$\pm$1.5$^{\uparrow}$ & 69.1$\pm$1.6$^{\uparrow}$ & 70.6$\pm$1.7$^{\Uparrow}$  & 73.3 \\
    \midrule
    STORM (ours)                 & 91.0$\pm$0.7$^{\uparrow}$ & 82.3$\pm$3.1$^{\Uparrow}$ & 51.8$\pm$2.1$^{\uparrow}$ & 86.8$\pm$1.2$^{\Uparrow}$ & 68.9$\pm$2.7 & 67.1$\pm$0.7$^{\Uparrow}$ & 74.7 \\
    \quad w/ binary rescaling & \textbf{92.5}$\pm$1.0$^{\Uparrow}$ & 15.3$\pm$24.7$^{\Downarrow}$ & 0.0$\pm$0.0$^{\Downarrow}$ & 83.9$\pm$2.7 & 69.4$\pm$1.9$^{\uparrow}$ & \textbf{67.4}$\pm$0.7$^{\Uparrow}$ & 54.8 \\
    \quad w/ 2 inner loops       & 89.6$\pm$1.3 & 84.2$\pm$2.7$^{\Uparrow}$ & 51.9$\pm$2.8$^{\uparrow}$ & \textbf{87.8}$\pm$1.7$^{\Uparrow}$ & 67.3$\pm$4.0 & 61.1$\pm$6.8 & 73.7 \\ % Last updated (minor): 2024.08.21
    \quad w/ 10 forward passes   & 92.1$\pm$0.7$^{\Uparrow}$ & 83.4$\pm$3.0$^{\Uparrow}$ & \textbf{53.2}$\pm$2.6$^{\Uparrow}$ & 87.3$\pm$1.6$^{\Uparrow}$ & \textbf{70.1}$\pm$2.3$^{\Uparrow}$ & 62.9$\pm$2.7 & \textbf{74.8} \\
    \midrule
    \quad w/o class separation   & 91.9$\pm$0.7$^{\Uparrow}$ & \textbf{84.7}$\pm$1.0$^{\Uparrow}$ & 51.0$\pm$2.6 & 87.1$\pm$1.9$^{\Uparrow}$ & 69.2$\pm$2.2$^{\uparrow}$ & 64.5$\pm$1.5 & 74.7 \\
    \quad w/o extra features     & 90.6$\pm$1.5 & 83.6$\pm$2.0$^{\Uparrow}$ & 50.2$\pm$2.0 & 86.7$\pm$1.6$^{\Uparrow}$ & 68.9$\pm$2.2 & 39.2$\pm$18.6$^{\downarrow}$ & 69.9 \\
    \quad w/o meta learning      & 86.5$\pm$1.9$^{\Downarrow}$ & 73.3$\pm$7.5$^{\Uparrow}$ & 48.1$\pm$4.7 & 85.7$\pm$2.5 & 68.7$\pm$0.9 & 65.0$\pm$0.9 & 71.2 \\
    \quad w/o meta loss rescaling & 87.6$\pm$2.5$^{\downarrow}$ & 65.5$\pm$4.8 & 48.8$\pm$3.3 & 84.6$\pm$1.0 & 69.2$\pm$2.5$^{\uparrow}$ & 66.3$\pm$1.1$^{\uparrow}$ & 70.3 \\
    \midrule\midrule
    \textit{Noisy labels} & \multicolumn{5}{c}{\textit{(uniform noise 40\%)}} & \\
    \midrule
    No rescaling                 & 81.8$\pm$3.7 & 48.6$\pm$3.7 & 41.6$\pm$3.2 & 81.1$\pm$0.8 & 67.6$\pm$1.8 & / & 64.1 \\
    \midrule
    STORM (ours)                 & 83.6$\pm$2.5 & 64.9$\pm$4.1$^{\Uparrow}$ & 42.8$\pm$2.9 & 81.6$\pm$0.9 & 68.8$\pm$0.5$^{\uparrow}$ & / & 68.3 \\
    \bottomrule
  \end{tabular}
  \caption{Model performance. ${\Uparrow}$/${\uparrow}$ and ${\Downarrow}$/${\downarrow}$ indicate significant ($p\ll 0.01$ vs. $p<0.05$) differences to \emph{no rescaling}.}
  \label{tab:results}
\end{table*}

\subsection{DST with TripPy}
\label{sec:dst:ssec:trippy}

We employ a triple copy strategy DST~\cite{heck-etal-2020-trippy}. The intuition is that values are either directly expressed by the user and can be extracted from context (\texttt{span}), values are mentioned by the system and indirectly referred to by the user and therefore can be retrieved from the system output (\informvalue{}), or values are coreferences to concepts that were mentioned earlier and can therefore be copied over (\refervalue{}). The special value \dontcarevalue{} is to represent user's indifference, \texttt{true} and \texttt{false} are for Boolean slots, and \texttt{none} is the empty slot. A slot gate decides which of these mechanisms or special values to use for filling a slot. Slot gates are implemented as individual classification heads on top of an encoder, analogous to Eq.~\ref{eq:theta}.
The total loss for one training example is a combination of
various classification losses.
Each of TripPy's various classification tasks requires their own set of labels. A labeling error occurring during any of the partial loss computations affects the total sample loss $\mathcal{L}_i$ and therefore the utility of that entire sample.

We focus on the most influential component for loss rescaling, the slot gates, which predict whether and how $S_i$ need to be filled. If a slot gate makes a faulty prediction, the predictions of \texttt{span} and \refervalue{} heads become irrelevant. Therefore, we rescale the sample losses as follows:
\begin{equation}
    \mathcal{L}_i^{'} 
    = 
    \sum^S_{s=1}\alpha \cdot \Omega_\omega(\vctr{f}_{s,i}) \cdot \mathcal{L}^{\gate}_{s,i} + \beta \cdot \mathcal{L}^{\spanup}_{s,i} + \gamma \cdot \mathcal{L}^{\referup}_{s,i}.
\end{equation}

% ================================
\section{Experiments}
\label{sec:experiments}
% ================================

\begin{figure*}[t]
	\centering
	\begin{minipage}{0.19\linewidth}
		\centering
		\includegraphics[width=\linewidth]{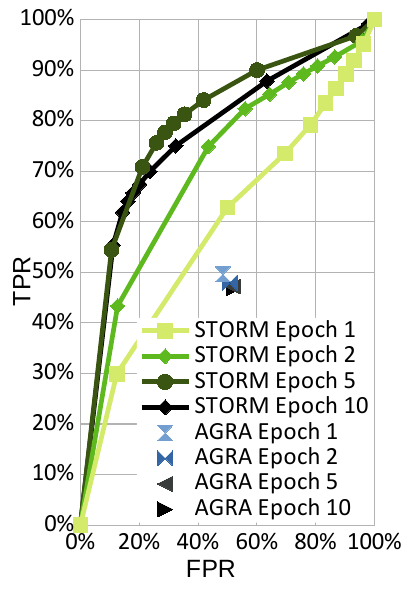}
		\caption*{(a) Rescaling ROC}
        \label{fig:roc}
	\end{minipage}
	\hfill
	\begin{minipage}{0.19\linewidth}
		\centering
		\includegraphics[width=\linewidth]{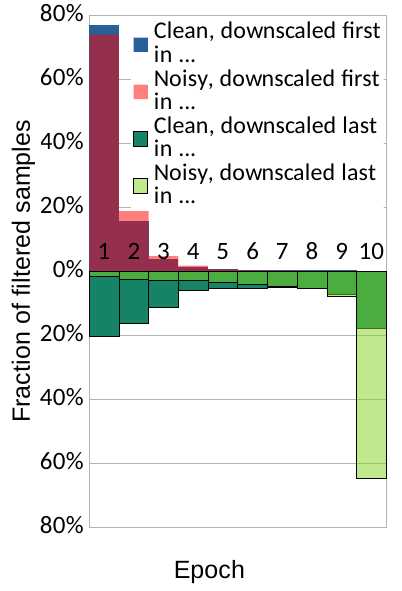}
		\caption*{(b) Filter timing}
        \label{fig:rescale_epochs}
	\end{minipage}
	\hfill
	\begin{minipage}{0.19\linewidth}
		\centering
		\includegraphics[width=1.0\linewidth]{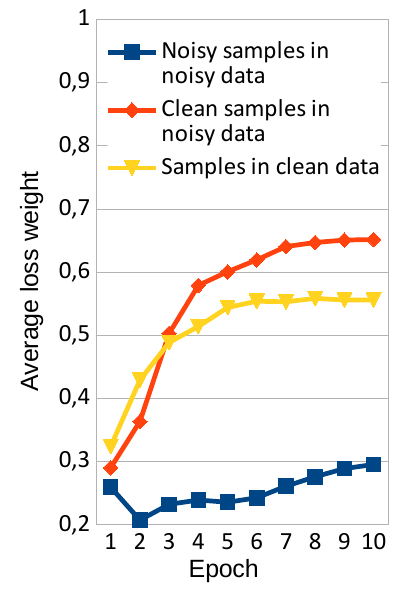}
        \caption*{(c) Weight progression}
        \label{fig:weight_progression}
	\end{minipage}
	\hfill
	\begin{minipage}{0.19\linewidth}
		\centering
		\includegraphics[width=\linewidth]{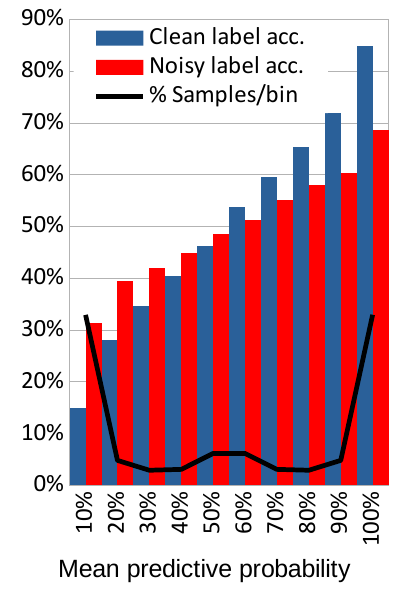}
		\caption*{(d) Model calibration}
        \label{fig:calibration}
	\end{minipage}
	\hfill
	\begin{minipage}{0.19\linewidth}
		\centering
		\includegraphics[width=\linewidth]{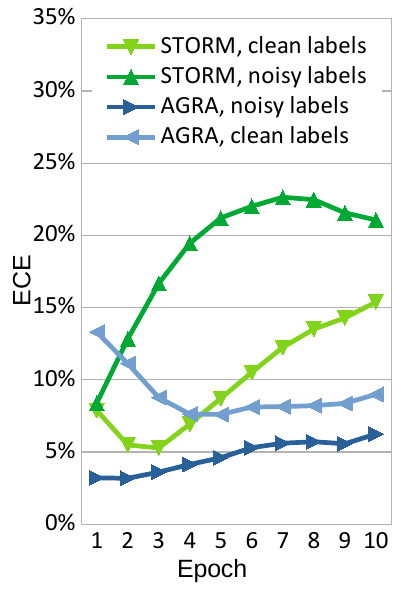}
		\caption*{(e) Calibration error}
        \label{fig:ece}
	\end{minipage}
    \caption{Rescaling analysis: (a) ROC curve for weights from 0.0 to 1.0, (b) Earliest and latest epochs for downscaling noisy and clean data, (c) Average weights per sample type across epochs; Model analysis: (d) Model calibration given noisy and clean samples, (e) Expected calibration error given noisy and clean labels. Reported are averages over all symmetric noise datasets.}
    \label{fig:ana}
\end{figure*}

\subsection{Datasets}

We train and evaluate on three types of NLP classification datasets. Youtube~\cite{7424299} and SMS~\cite{10.1145/2034691.2034742} are spam detection benchmarks. We encode the samples for these two datasets with fixed TF-IDF vectors. MRPC, CoLA and RTE are members of the GLUE benchmark~\cite{wang-etal-2018-glue} and cover the tasks of paraphrase, linguistic acceptability and textual entailment detection, respectively. 
%The samples of these datasets are encoded with trainable contextual representations of our classifier's transformer based encoder.
We introduce 10\% to 40\% random label noise into the training portions of these datasets to simulate noisy labels.
MultiWOZ 2.4~\cite{ye-etal-2022-multiwoz} is a task oriented dialogue modeling benchmark that takes training set of MultiWOZ 2.1, but provides rigorously cleaned validation and test sets. Its noise stems mainly from inter-annotator inconsistencies and systematic intra-annotator mistakes w.r.t. to the labeling instructions, making labels frequently ambiguous.
The datasets we selected vary in size, balance, sparsity and complexity, and thus provide individual challenges to our approach.
The YouTube, SMS, CoLA, MRPC, and RTE datasets each contain two classes and include 30\% symmetric noise. The datasets consist of 1.6K, 4.5K, 8.5K, 6.7K, and 2.5K samples, respectively. Among these, only the YouTube dataset does not exhibit class imbalance. The MultiWOZ dialogue dataset spans 7 domains, contains 20-40\% real noise, has 56.8K samples, and is affected by class imbalance.

\subsection{Evaluation}

As performance metrics for the model, we use accuracy for balanced data, F1 score for imbalanced data, Matthew's correlation for the CoLA benchmark and joint goal accuracy (JGA) for DST on MultiWOZ. JGA is the ratio of dialogue turns for which all slots were filled with the correct value according to the dialogue state labels (including the \texttt{none} value).
We run each experiment 10 (for MultiWOZ 5) times with random seeds. We report averages, standard deviation and statistical significance of the results.
We perform model selection given the validation sets and test on the test sets. For GLUE benchmarks, no test sets are available, therefore we use 2-fold cross-validation using the validation sets.
We perform an ablation study on data with a uniform noise ratio of 30\% and on data with real noise.

\subsection{Training and Inference}

We initialize $\text{Enc}(\cdot)$ with RoBERTa-base~\cite{liu2019robertarobustlyoptimizedbert}. All tasks are trained with cross-entropy loss using the Adam optimizer~\cite{kingma2017adammethodstochasticoptimization}. During backpropagation, we also pass through $\text{Enc}(\cdot)$. Optimal learning rates are determined via grid search on the original clean datasets. MultiWOZ experiments are an exception due to the lack of a clean training dataset. Learning rates are constant except for MultiWOZ, where we employ a linear schedule with 10\% warmup. Maximum epochs are 10 with early stopping based on validation performance. Batch sizes $B$ are 48 for MultiWOZ and 32 for the other datasets. The dropout rate for the transformer encoder is 10\%. Since the experiments with TF-IDF features do not utilize an encoder, we directly dropout the features at the same rate.
To isolate the effect of rescaling, we maintain the same setup and dataset specific hyperparameters unrelated to rescaling for all experiments. Any changes in performance can therefore be directly attributed to the rescaling.
We performed a simple hyperparameter search for meta learning across all corpora so as to avoid task-specific tuning. We found no statistically significant improvements with alternative hyperparameter values.
We found that using the full set of features for $\Omega$ leads to most consistent results, compared to using subsets.
To minimize computational overhead,  
we set $G=3$ and pass through the inner loop of meta learning once per training step. We set $m_\text{size} = B$.
As baseline, we use AGRA with cross-entropy as comparison loss and without weighted sampling~\cite{Sedova_2023}.

\begin{figure*}[t]
\centering
\begin{minipage}[t]{0.68\textwidth}
    \centering
    \small
    \vspace{0pt} % Align at the top of the minipage
    \begin{tabular}{p{0.5cm} p{4.2cm} p{4.2cm} p{1.6cm}}
    \toprule
    \# & \textbf{User at $t+1$} & \textbf{System at $t$} & \textbf{Label} \\ \midrule
    1 & I am leaving on Friday. What is the cost please? & What day are you taking the train? & none (wrong) \\ \midrule
    2 & It does not need internet included & Do you have any additional preferences? & dontcare (ambiguous) \\ \midrule
    3 & It does not matter. I'm looking for a nice museum. & What area of town were you looking to visit? & dontcare (hard) \\ \midrule
    4 & Yes, I need some information on Rosa's Bed and Breakfast. & Is there anything else? & Rosa's ... (easy) \\ \bottomrule
    \end{tabular}
\end{minipage}
\hfill
\begin{minipage}[t]{0.30\textwidth}
    \centering
    \vspace{0pt} 
    \includegraphics[width=.95\linewidth]{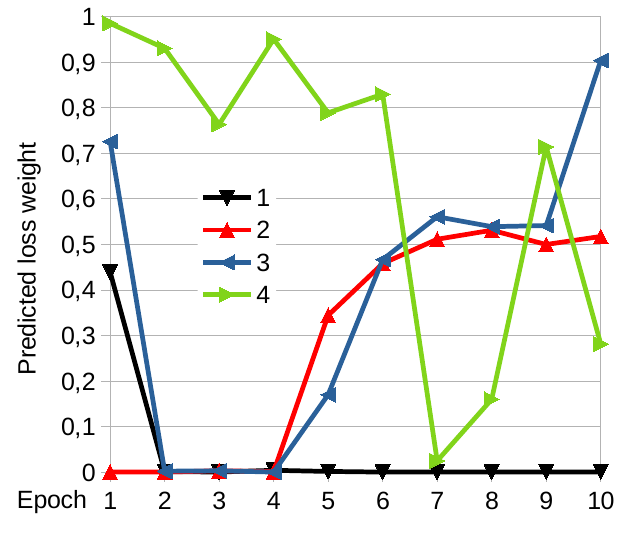}
\end{minipage}
\caption{\#1: A wrongly labeled sample is discarded consistently after initial consideration in the first epoch. \#2: An ambiguous sample is initially ignored, then gradually upscaled. \#3: A clean but hard sample sees its loss weight increased steadily from mid-training. \#4: A clean and easy sample has its high initial loss weight reduced due to decreasing informativeness.}
\label{fig:qualitative}
\end{figure*}

\subsection{Clean vs. Noisy Labels}

An experiment often neglected is testing the effect of loss rescaling on clean data. We observed that STORM does not significantly affect training on non-noisy data, i.e., performance is not diminished (Tab.~\ref{tab:results}). For comparison, AGRA at times diminishes performance on clean data, likely due to over-filtering.
%(more on that in Section~\ref{sec:experiments:ssec:binary}).
Our approach merely rescales losses and can therefore utilize all samples. Fig.~\ref{fig:ana}c shows that loss weights increase over time, leading to a higher impact of individual samples later in training.

STORM significantly outperforms training on noisy data in all our experiments (Tab.~\ref{tab:results}). The largest gain is observed on the TF-IDF encoded SMS dataset, where STORM improves performance by 19\%-21\% absolute. Transformer-encoder based models likewise perform considerably better, especially in the challenging MultiWOZ DST task. While AGRA improves some models, it tends to harm the DST model compared to simply training on noisy data.

\subsection{Rescaling vs. Removal}
\label{sec:experiments:ssec:binary}

Since AGRA performs binary loss rescaling, we implemented a binary version of STORM for comparison. Our alternative approach underperforms compared to AGRA and standard STORM (Tab.~\ref{tab:results}, ``STORM w/ binary rescaling''). Although binary STORM achieves good results on some benchmarks, e.g., Youtube, it fails on others such as SMS and CoLA.

Hard filtering of samples risks losing critical information, preventing the model from escaping unfavorable parameter regions. If this happens early in the training, $\Theta$'s signals to $\Omega$ become uninformative, breaking their feedback loop. Hard filtering data changes its distribution, which violates the i.i.d. assumption in sampling batch $\mathfrak{B}_\text{val}$ for the meta update. This causes an imbalance that is hard to recover from. AGRA avoids this issue, likely because it is gradient based and not error based, and uses a threshold on a similarity measure instead of learning a rescaling or filtering function.

\subsection{Noisy vs. Clean Validation Signal}

Sampling $\mathfrak{B}_\text{val}$ from a clean validation set,
contrary to expectations, did not generally improve rescaling quality or model performance. In fact, meta loss rescaling in the outer loop of meta learning was detrimental. Tab.~\ref{tab:results} reports model performance when meta learning $\Omega$ without meta loss rescaling and without considering $\nabla_\text{outer} \omega$ (Tab.~\ref{tab:results}, ``Meta learning w/ clean val.'').

Using clean validation data, We observed a significant improvement only in DST, but not in simpler NLP tasks. This is due to the nature of the dataset noise. For uniform noise, clean validation data offers no extra benefit beyond what we can learn from noisy data, since randomly noised samples do not exhibit a pattern. Consequently, we did not see statistically significant differences in performance between using clean and noisy validation signals. In the DST experiments on the MultiWOZ data, which contains real noise, the clean data proves helpful to handle non-random biases from human labeling.

\subsection{Ablation}

\paragraph{Do we need meta learning?} Meta learning is essential for learning a loss rescaler from noisy labels. Without it, performance remains similar to conventional training on noisy data (Tab.~\ref{tab:results}, ``STORM w/o meta learning''), mainly due to overfitting to noise. As $\Theta$ overfits, without meta learning $\Omega$ cannot distinguish between good and noisy samples effectively due to their features $\vctr{f}_i$ being increasingly indistinguishable, causing over-filtering of correct samples.

\paragraph{Do we need meta loss rescaling?} Without combining $\nabla_\text{meta} \omega$ and $\nabla_\text{outer} \omega$ for updating $\omega$  (Tab.~\ref{tab:results}, ``STORM w/o meta loss rescaling''), the loss rescaler remains conservative in its estimates, leading to only a small gap between weights for clean and noisy samples. Meta loss rescaling ensures a much clearer differentiation (Fig.~\ref{fig:ana}c). Without the complementary goals of inner and outer loop w.r.t. updating $\omega$, optimization would simply minimize the loss of the noisy validation data, leading to overfitting $\Theta$ and uninformative $\vctr{f}_i$. 

\paragraph{Do we need more than loss as features?} While STORM can work with only sample loss as a feature ($\vctr{f}_i = \ell_i$) on symmetric noise benchmarks (Tab.~\ref{tab:results}, ``STORM w/o extra features''), it underperforms on real noise. The additional features we compute are complementary to the sample loss and evidently help identify non-random biases in the data.

\paragraph{Do we need class dependent rescaling?} Target class dependent $f_i$ and $\Omega$ had no impact on simpler benchmarks with moderate or no class imbalance (Tab.~\ref{tab:results}, ``STORM w/o class separation''). In the DST use case, however, class dependent rescaling was critical to achieve good performance. In MultiWOZ, class imbalance is extreme, leading to different class statistics. Joint rescaling would categorically downscale underrepresented classes, rather than individual outliers.

\paragraph{Is ``more'' better?} More inner loop traversals before an outer loop update did not improve performance, but considerably increased training time and memory consumption (Tab.~\ref{tab:results}, ``STORM w/ 2 inner loops''). More gradient-free forward passes for richer statistics in $\vctr{f}_i$ slightly improved performance but the benefit was disproportionally small compared to the increased cost (Tab.~\ref{tab:results}, ``STORM w/ 10 forward passes'').

\subsection{Use Case: DST with TripPy}

Our use case is representative of dialogue modeling tasks heavily affected noisy data. Applying STORM to train TripPy DST for MultiWOZ with its estimated 20\%-40\% noise improved performance by over 2\% JGA absolute.

STORM shows slower training convergence but higher peak performance, indicating it avoids overfitting. Regular training, however, tends to overfit to data noise. Our baseline, AGRA, does not outperform regular training, possibly because it only considers classification head gradients due to memory constraints and does not distinguish target classes, unlike STORM.

Our approach consistently improves performance across all classes in MultiWOZ (+1\% F1, +4\% recall), particularly \dontcarevalue{} (+5\% F1, +8\% recall), an inconsistently labeled class, and \refervalue{} (+1\% F1, +9\% recall), the most challenging class, both being underrepresented.

\subsection{Analysis}

\paragraph{STORM keeps improving}

As meta learning progresses, STORM's loss rescaling function gradually improves in identifying noisy samples. Fig.~\ref{fig:ana}a plots the ROC curve of $\Omega$ averaged over all benchmarks. STORM continues learning from $\Theta$ throughout training, unlike our baseline, which maintains stable true/false positive rates. %~\cite{Sedova_2023}.

\paragraph{STORM downscales noise consistently}

Fig.~\ref{fig:ana}b shows that most downscaled samples, clean and noisy, are downscaled early on in training. The difference in downscaling behavior w.r.t. clean and noisy samples emerges as training progresses. Clean samples that were (wrongly) downscaled early in training stop being downscaled in later epochs. In contrast, noisy samples remain consistently downscaled until the end of training. A notable number of clean samples are downscaled throughout training, typically being ambiguous cases or from underrepresented groups of samples.

\paragraph{STORM creates a learning schedule}

STORM tends to (1) increase the gap between loss weights for clean and noisy samples over time, (2) downscale noisy samples faster than upscale clean ones, and (3) increase average loss across all samples (Fig.~\ref{fig:ana}c). These indicators suggest a learning schedule where noise is discarded early, and challenging clean samples are increasingly considered. Some previously discarded samples, likely ambiguous or hard cases, are reconsidered later in training.

\paragraph{STORM prevents overfitting}

Figures~\ref{fig:ana}d-e show how STORM reduces $\Theta$'s tendency to overfit noise. The calibration histogram indicates that a model trained with STORM maintains good calibration for clean sample labels but not for noisy labels, evidencing a resistance to overfitting. Fig.~\ref{fig:ana}e plots the expected calibration error (ECE) and shows that STORM avoids overfitting to noisy samples, unlike our baseline.

\paragraph{STORM reveals patterns} Our qualitative analysis reveals patterns based on a sample's correctness or difficulty. Exemplarily, Fig.~\ref{fig:qualitative} illustrates STORM's behavior on MultiWOZ. 

\subsection{Limitations}

We observed that the potential risk of confirmation bias in applying STORM using noisy validation data is heightened by two factors. First, samples of a particular class that are consistently labeled incorrectly have a higher risk to be wrongly down- or upscaled due to the training gradually learning to consider the wrong label(s) to be correct. Second, severely under-represented classes have a higher risk of being wrongly down- or upscaled if a particular rescaling trend is established for these samples early in the training and then reinforced throughout the remainder of the training. Both these risks we observed for the \dontcarevalue{} class in our use case for DST on MultiWOZ data, which is known to be a particularly difficult labeling task.

% ================================
\section{Conclusion}
\label{sec:conclusion}
% ================================

In conclusion, our self-taught on-the-fly meta loss rescaling demonstrates efficiency in learning from noisy labels, without the need for clean validation data. Our novel meta-learning scheme dynamically rescales losses in a flexible and computationally efficient approach as it leverages sample losses and prediction probabilities from the model being trained. STORM is robust towards class imbalances and different noise types, as shown by extensive empirical evaluation across various NLP tasks, including dialogue modeling. Remarkably, STORM consistently downscales noisy samples, develops an effective learning schedule, and prevents overfitting by ensuring good calibration for clean sample labels. We see limitations in applying STORM to extremely noisy data, where the clean samples are outnumbered, but consider this a prospective challenge for future work.

\section{Acknowledgments}

This work was made possible through the support
of the Alexander von Humboldt Foundation, provided within the Sofja Kovalevskaja Award, the
European Research Council (ERC) under the Horizon 2020 research and innovation program (Grant
No. STG2018 804636), and the Ministry of Culture and Science of North Rhine-Westphalia within
the Lamarr Fellow Network. Computational resources were provided by the Centre for Information and Media Technology at Heinrich Heine University Düsseldorf, and Google Cloud.

\bibliography{aaai25}

\begin{thebibliography}{39}
\providecommand{\natexlab}[1]{#1}

\bibitem[{Alberto, Lochter, and Almeida(2015)}]{7424299}
Alberto, T.~C.; Lochter, J.~V.; and Almeida, T.~A. 2015.
\newblock TubeSpam: Comment Spam Filtering on YouTube.
\newblock In \emph{2015 IEEE 14th International Conference on Machine Learning and Applications (ICMLA)}, 138--143.

\bibitem[{Almeida, Hidalgo, and Yamakami(2011)}]{10.1145/2034691.2034742}
Almeida, T.~A.; Hidalgo, J. M.~G.; and Yamakami, A. 2011.
\newblock Contributions to the study of SMS spam filtering: new collection and results.
\newblock In \emph{Proceedings of the 11th ACM Symposium on Document Engineering}, DocEng '11, 259–262. New York, NY, USA: Association for Computing Machinery.
\newblock ISBN 9781450308632.

\bibitem[{Balaraman, Sheikhalishahi, and Magnini(2021)}]{balaraman-etal-2021-recent}
Balaraman, V.; Sheikhalishahi, S.; and Magnini, B. 2021.
\newblock Recent Neural Methods on Dialogue State Tracking for Task-Oriented Dialogue Systems: A Survey.
\newblock In Li, H.; Levow, G.-A.; Yu, Z.; Gupta, C.; Sisman, B.; Cai, S.; Vandyke, D.; Dethlefs, N.; Wu, Y.; and Li, J.~J., eds., \emph{Proceedings of the 22nd Annual Meeting of the Special Interest Group on Discourse and Dialogue}, 239--251. Singapore and Online: Association for Computational Linguistics.

\bibitem[{Budzianowski et~al.(2018)Budzianowski, Wen, Tseng, Casanueva, Ultes, Ramadan, and Ga{\v{s}}i{\'c}}]{budzianowski-etal-2018-multiwoz}
Budzianowski, P.; Wen, T.-H.; Tseng, B.-H.; Casanueva, I.; Ultes, S.; Ramadan, O.; and Ga{\v{s}}i{\'c}, M. 2018.
\newblock {M}ulti{WOZ} - A Large-Scale Multi-Domain {W}izard-of-{O}z Dataset for Task-Oriented Dialogue Modelling.
\newblock In Riloff, E.; Chiang, D.; Hockenmaier, J.; and Tsujii, J., eds., \emph{Proceedings of the 2018 Conference on Empirical Methods in Natural Language Processing}, 5016--5026. Brussels, Belgium: Association for Computational Linguistics.

\bibitem[{Chang, Learned{-}Miller, and McCallum(2017)}]{chang2018activebiastrainingaccurate}
Chang, H.; Learned{-}Miller, E.~G.; and McCallum, A. 2017.
\newblock Active Bias: Training More Accurate Neural Networks by Emphasizing High Variance Samples.
\newblock In Guyon, I.; von Luxburg, U.; Bengio, S.; Wallach, H.~M.; Fergus, R.; Vishwanathan, S. V.~N.; and Garnett, R., eds., \emph{Advances in Neural Information Processing Systems 30: Annual Conference on Neural Information Processing Systems 2017, December 4-9, 2017, Long Beach, CA, {USA}}, 1002--1012.

\bibitem[{Chen et~al.(2019)Chen, Liao, Chen, and Zhang}]{chen2019understandingutilizingdeepneural}
Chen, P.; Liao, B.; Chen, G.; and Zhang, S. 2019.
\newblock Understanding and Utilizing Deep Neural Networks Trained with Noisy Labels.
\newblock In Chaudhuri, K.; and Salakhutdinov, R., eds., \emph{Proceedings of the 36th International Conference on Machine Learning, {ICML} 2019, 9-15 June 2019, Long Beach, California, {USA}}, volume~97 of \emph{Proceedings of Machine Learning Research}, 1062--1070. {PMLR}.

\bibitem[{Dehghani et~al.(2017)Dehghani, Severyn, Rothe, and Kamps}]{dehghani2017learninglearnweaksupervision}
Dehghani, M.; Severyn, A.; Rothe, S.; and Kamps, J. 2017.
\newblock Learning to Learn from Weak Supervision by Full Supervision.
\newblock \emph{CoRR}, abs/1711.11383.

\bibitem[{Finn, Abbeel, and Levine(2017)}]{finn2017modelagnosticmetalearningfastadaptation}
Finn, C.; Abbeel, P.; and Levine, S. 2017.
\newblock Model-Agnostic Meta-Learning for Fast Adaptation of Deep Networks.
\newblock In Precup, D.; and Teh, Y.~W., eds., \emph{Proceedings of the 34th International Conference on Machine Learning, {ICML} 2017, Sydney, NSW, Australia, 6-11 August 2017}, volume~70 of \emph{Proceedings of Machine Learning Research}, 1126--1135. {PMLR}.

\bibitem[{Frenay and Verleysen(2014)}]{6685834}
Frenay, B.; and Verleysen, M. 2014.
\newblock Classification in the Presence of Label Noise: A Survey.
\newblock \emph{IEEE Transactions on Neural Networks and Learning Systems}, 25(5): 845--869.

\bibitem[{Han et~al.(2018)Han, Yao, Yu, Niu, Xu, Hu, Tsang, and Sugiyama}]{han2018coteachingrobusttrainingdeep}
Han, B.; Yao, Q.; Yu, X.; Niu, G.; Xu, M.; Hu, W.; Tsang, I.~W.; and Sugiyama, M. 2018.
\newblock Co-teaching: Robust training of deep neural networks with extremely noisy labels.
\newblock In Bengio, S.; Wallach, H.~M.; Larochelle, H.; Grauman, K.; Cesa{-}Bianchi, N.; and Garnett, R., eds., \emph{Advances in Neural Information Processing Systems 31: Annual Conference on Neural Information Processing Systems 2018, NeurIPS 2018, December 3-8, 2018, Montr{\'{e}}al, Canada}, 8536--8546.

\bibitem[{Heck et~al.(2020)Heck, van Niekerk, Lubis, Geishauser, Lin, Moresi, and Gasic}]{heck-etal-2020-trippy}
Heck, M.; van Niekerk, C.; Lubis, N.; Geishauser, C.; Lin, H.-C.; Moresi, M.; and Gasic, M. 2020.
\newblock {T}rip{P}y: A Triple Copy Strategy for Value Independent Neural Dialog State Tracking.
\newblock In Pietquin, O.; Muresan, S.; Chen, V.; Kennington, C.; Vandyke, D.; Dethlefs, N.; Inoue, K.; Ekstedt, E.; and Ultes, S., eds., \emph{Proceedings of the 21th Annual Meeting of the Special Interest Group on Discourse and Dialogue}, 35--44. 1st virtual meeting: Association for Computational Linguistics.

\bibitem[{Huang et~al.(2019)Huang, Qu, Jia, and Zhao}]{9008796}
Huang, J.; Qu, L.; Jia, R.; and Zhao, B. 2019.
\newblock O2U-Net: A Simple Noisy Label Detection Approach for Deep Neural Networks.
\newblock In \emph{2019 IEEE/CVF International Conference on Computer Vision (ICCV)}, 3325--3333.

\bibitem[{Inman and Jr(1989)}]{doi:10.1080/03610928908830127}
Inman, H.~F.; and Jr, E. L.~B. 1989.
\newblock The overlapping coefficient as a measure of agreement between probability distributions and point estimation of the overlap of two normal densities.
\newblock \emph{Communications in Statistics - Theory and Methods}, 18(10): 3851--3874.

\bibitem[{Ioffe and Szegedy(2015)}]{ioffe2015batchnormalizationacceleratingdeep}
Ioffe, S.; and Szegedy, C. 2015.
\newblock Batch Normalization: Accelerating Deep Network Training by Reducing Internal Covariate Shift.
\newblock In Bach, F.~R.; and Blei, D.~M., eds., \emph{Proceedings of the 32nd International Conference on Machine Learning, {ICML} 2015, Lille, France, 6-11 July 2015}, volume~37 of \emph{{JMLR} Workshop and Conference Proceedings}, 448--456. JMLR.org.

\bibitem[{Jiang et~al.(2018)Jiang, Zhou, Leung, Li, and Fei{-}Fei}]{jiang2018mentornetlearningdatadrivencurriculum}
Jiang, L.; Zhou, Z.; Leung, T.; Li, L.; and Fei{-}Fei, L. 2018.
\newblock MentorNet: Learning Data-Driven Curriculum for Very Deep Neural Networks on Corrupted Labels.
\newblock In Dy, J.~G.; and Krause, A., eds., \emph{Proceedings of the 35th International Conference on Machine Learning, {ICML} 2018, Stockholmsm{\"{a}}ssan, Stockholm, Sweden, July 10-15, 2018}, volume~80 of \emph{Proceedings of Machine Learning Research}, 2309--2318. {PMLR}.

\bibitem[{Kelley(1984)}]{10.1145/357417.357420}
Kelley, J.~F. 1984.
\newblock An iterative design methodology for user-friendly natural language office information applications.
\newblock \emph{ACM Trans. Inf. Syst.}, 2(1): 26–41.

\bibitem[{Kingma and Ba(2015)}]{kingma2017adammethodstochasticoptimization}
Kingma, D.~P.; and Ba, J. 2015.
\newblock Adam: {A} Method for Stochastic Optimization.
\newblock In Bengio, Y.; and LeCun, Y., eds., \emph{3rd International Conference on Learning Representations, {ICLR} 2015, San Diego, CA, USA, May 7-9, 2015, Conference Track Proceedings}.

\bibitem[{Li et~al.(2019)Li, Wong, Zhao, and Kankanhalli}]{li2019learninglearnnoisylabeled}
Li, J.; Wong, Y.; Zhao, Q.; and Kankanhalli, M.~S. 2019.
\newblock Learning to Learn From Noisy Labeled Data.
\newblock In \emph{{IEEE} Conference on Computer Vision and Pattern Recognition, {CVPR} 2019, Long Beach, CA, USA, June 16-20, 2019}, 5051--5059. Computer Vision Foundation / {IEEE}.

\bibitem[{Li, Soltanolkotabi, and Oymak(2020)}]{li2019gradientdescentearlystopping}
Li, M.; Soltanolkotabi, M.; and Oymak, S. 2020.
\newblock Gradient Descent with Early Stopping is Provably Robust to Label Noise for Overparameterized Neural Networks.
\newblock In Chiappa, S.; and Calandra, R., eds., \emph{The 23rd International Conference on Artificial Intelligence and Statistics, {AISTATS} 2020, 26-28 August 2020, Online [Palermo, Sicily, Italy]}, volume 108 of \emph{Proceedings of Machine Learning Research}, 4313--4324. {PMLR}.

\bibitem[{Liu and Tao(2016)}]{Liu_2016}
Liu, T.; and Tao, D. 2016.
\newblock Classification with Noisy Labels by Importance Reweighting.
\newblock \emph{IEEE Transactions on Pattern Analysis and Machine Intelligence}, 38(3): 447–461.

\bibitem[{Liu et~al.(2019)Liu, Ott, Goyal, Du, Joshi, Chen, Levy, Lewis, Zettlemoyer, and Stoyanov}]{liu2019robertarobustlyoptimizedbert}
Liu, Y.; Ott, M.; Goyal, N.; Du, J.; Joshi, M.; Chen, D.; Levy, O.; Lewis, M.; Zettlemoyer, L.; and Stoyanov, V. 2019.
\newblock RoBERTa: {A} Robustly Optimized {BERT} Pretraining Approach.
\newblock \emph{CoRR}, abs/1907.11692.

\bibitem[{Manning(2011)}]{10.1007/978-3-642-19400-9_14}
Manning, C.~D. 2011.
\newblock Part-of-Speech Tagging from 97{\%} to 100{\%}: Is It Time for Some Linguistics?
\newblock In Gelbukh, A.~F., ed., \emph{Computational Linguistics and Intelligent Text Processing}, 171--189. Berlin, Heidelberg: Springer Berlin Heidelberg.

\bibitem[{Marcus, Santorini, and Marcinkiewicz(1993)}]{marcus-etal-1993-building}
Marcus, M.~P.; Santorini, B.; and Marcinkiewicz, M.~A. 1993.
\newblock Building a Large Annotated Corpus of {E}nglish: The {P}enn {T}reebank.
\newblock \emph{Computational Linguistics}, 19(2): 313--330.

\bibitem[{Patrini et~al.(2017)Patrini, Rozza, Menon, Nock, and Qu}]{patrini2017makingdeepneuralnetworks}
Patrini, G.; Rozza, A.; Menon, A.~K.; Nock, R.; and Qu, L. 2017.
\newblock Making Deep Neural Networks Robust to Label Noise: {A} Loss Correction Approach.
\newblock In \emph{2017 {IEEE} Conference on Computer Vision and Pattern Recognition, {CVPR} 2017, Honolulu, HI, USA, July 21-26, 2017}, 2233--2241. {IEEE} Computer Society.

\bibitem[{Reed et~al.(2015)Reed, Lee, Anguelov, Szegedy, Erhan, and Rabinovich}]{reed2015trainingdeepneuralnetworks}
Reed, S.~E.; Lee, H.; Anguelov, D.; Szegedy, C.; Erhan, D.; and Rabinovich, A. 2015.
\newblock Training Deep Neural Networks on Noisy Labels with Bootstrapping.
\newblock In Bengio, Y.; and LeCun, Y., eds., \emph{3rd International Conference on Learning Representations, {ICLR} 2015, San Diego, CA, USA, May 7-9, 2015, Workshop Track Proceedings}.

\bibitem[{Ren et~al.(2018)Ren, Zeng, Yang, and Urtasun}]{ren2019learningreweightexamplesrobust}
Ren, M.; Zeng, W.; Yang, B.; and Urtasun, R. 2018.
\newblock Learning to Reweight Examples for Robust Deep Learning.
\newblock In Dy, J.~G.; and Krause, A., eds., \emph{Proceedings of the 35th International Conference on Machine Learning, {ICML} 2018, Stockholmsm{\"{a}}ssan, Stockholm, Sweden, July 10-15, 2018}, volume~80 of \emph{Proceedings of Machine Learning Research}, 4331--4340. {PMLR}.

\bibitem[{Sedova, Zellinger, and Roth(2023)}]{Sedova_2023}
Sedova, A.; Zellinger, L.; and Roth, B. 2023.
\newblock \emph{Learning with Noisy Labels by Adaptive Gradient-Based Outlier Removal}, 237–253.
\newblock Springer Nature Switzerland.
\newblock ISBN 9783031434129.

\bibitem[{Shen et~al.(2024)Shen, Yao, Huang, Wang, Zhang, Wang, Yu, and Liu}]{shen2024protosimi}
Shen, J.; Yao, Y.; Huang, S.; Wang, Z.; Zhang, J.; Wang, R.; Yu, J.; and Liu, T. 2024.
\newblock ProtoSimi: label correction for fine-grained visual categorization.
\newblock \emph{Machine Learning}, 113(4): 1903--1920.

\bibitem[{Shen and Sanghavi(2019)}]{shen2019learningbadtrainingdata}
Shen, Y.; and Sanghavi, S. 2019.
\newblock Learning with Bad Training Data via Iterative Trimmed Loss Minimization.
\newblock In Chaudhuri, K.; and Salakhutdinov, R., eds., \emph{Proceedings of the 36th International Conference on Machine Learning, {ICML} 2019, 9-15 June 2019, Long Beach, California, {USA}}, volume~97 of \emph{Proceedings of Machine Learning Research}, 5739--5748. {PMLR}.

\bibitem[{Shu et~al.(2019)Shu, Xie, Yi, Zhao, Zhou, Xu, and Meng}]{shu2019metaweightnetlearningexplicitmapping}
Shu, J.; Xie, Q.; Yi, L.; Zhao, Q.; Zhou, S.; Xu, Z.; and Meng, D. 2019.
\newblock Meta-Weight-Net: Learning an Explicit Mapping For Sample Weighting.
\newblock In Wallach, H.~M.; Larochelle, H.; Beygelzimer, A.; d'Alch{\'{e}}{-}Buc, F.; Fox, E.~B.; and Garnett, R., eds., \emph{Advances in Neural Information Processing Systems 32: Annual Conference on Neural Information Processing Systems 2019, NeurIPS 2019, December 8-14, 2019, Vancouver, BC, Canada}, 1917--1928.

\bibitem[{Stolcke et~al.(2000)Stolcke, Ries, Coccaro, Shriberg, Bates, Jurafsky, Taylor, Martin, Van Ess-Dykema, and Meteer}]{stolcke-etal-2000-dialogue}
Stolcke, A.; Ries, K.; Coccaro, N.; Shriberg, E.; Bates, R.; Jurafsky, D.; Taylor, P.; Martin, R.; Van Ess-Dykema, C.; and Meteer, M. 2000.
\newblock Dialogue act modeling for automatic tagging and recognition of conversational speech.
\newblock \emph{Computational Linguistics}, 26(3): 339--374.

\bibitem[{Traum(2000)}]{10.1093/jos/17.1.7}
Traum, D.~R. 2000.
\newblock {20 Questions on Dialogue Act Taxonomies}.
\newblock \emph{Journal of Semantics}, 17(1): 7--30.

\bibitem[{Wang et~al.(2018)Wang, Singh, Michael, Hill, Levy, and Bowman}]{wang-etal-2018-glue}
Wang, A.; Singh, A.; Michael, J.; Hill, F.; Levy, O.; and Bowman, S. 2018.
\newblock {GLUE}: A Multi-Task Benchmark and Analysis Platform for Natural Language Understanding.
\newblock In Linzen, T.; Chrupa{\l}a, G.; and Alishahi, A., eds., \emph{Proceedings of the 2018 {EMNLP} Workshop {B}lackbox{NLP}: Analyzing and Interpreting Neural Networks for {NLP}}, 353--355. Brussels, Belgium: Association for Computational Linguistics.

\bibitem[{Wang, Liu, and Tao(2018)}]{7929355}
Wang, R.; Liu, T.; and Tao, D. 2018.
\newblock Multiclass Learning With Partially Corrupted Labels.
\newblock \emph{IEEE Transactions on Neural Networks and Learning Systems}, 29(6): 2568--2580.

\bibitem[{Wen et~al.(2017)Wen, Vandyke, Mrk{\v{s}}i{\'c}, Ga{\v{s}}i{\'c}, Rojas-Barahona, Su, Ultes, and Young}]{wen-etal-2017-network}
Wen, T.-H.; Vandyke, D.; Mrk{\v{s}}i{\'c}, N.; Ga{\v{s}}i{\'c}, M.; Rojas-Barahona, L.~M.; Su, P.-H.; Ultes, S.; and Young, S. 2017.
\newblock A Network-based End-to-End Trainable Task-oriented Dialogue System.
\newblock In Lapata, M.; Blunsom, P.; and Koller, A., eds., \emph{Proceedings of the 15th Conference of the {E}uropean Chapter of the Association for Computational Linguistics: Volume 1, Long Papers}, 438--449. Valencia, Spain: Association for Computational Linguistics.

\bibitem[{Yang et~al.(2019)Yang, Yao, Han, and Niu}]{yao2020searchingexploitmemorizationeffect}
Yang, H.; Yao, Q.; Han, B.; and Niu, G. 2019.
\newblock Searching to Exploit Memorization Effect in Learning from Corrupted Labels.
\newblock \emph{CoRR}, abs/1911.02377.

\bibitem[{Ye, Manotumruksa, and Yilmaz(2022)}]{ye-etal-2022-multiwoz}
Ye, F.; Manotumruksa, J.; and Yilmaz, E. 2022.
\newblock {M}ulti{WOZ} 2.4: A Multi-Domain Task-Oriented Dialogue Dataset with Essential Annotation Corrections to Improve State Tracking Evaluation.
\newblock In Lemon, O.; Hakkani-Tur, D.; Li, J.~J.; Ashrafzadeh, A.; Garcia, D.~H.; Alikhani, M.; Vandyke, D.; and Du{\v{s}}ek, O., eds., \emph{Proceedings of the 23rd Annual Meeting of the Special Interest Group on Discourse and Dialogue}, 351--360. Edinburgh, UK: Association for Computational Linguistics.

\bibitem[{Young et~al.(2010)Young, Gašić, Keizer, Mairesse, Schatzmann, Thomson, and Yu}]{YOUNG2010150}
Young, S.; Gašić, M.; Keizer, S.; Mairesse, F.; Schatzmann, J.; Thomson, B.; and Yu, K. 2010.
\newblock The Hidden Information State model: A practical framework for POMDP-based spoken dialogue management.
\newblock \emph{Computer Speech \& Language}, 24(2): 150--174.

\bibitem[{Zhang, Xing, and Liu(2021)}]{9577967}
Zhang, H.; Xing, X.; and Liu, L. 2021.
\newblock DualGraph: A graph-based method for reasoning about label noise.
\newblock In \emph{2021 IEEE/CVF Conference on Computer Vision and Pattern Recognition (CVPR)}, 9649--9658.

\end{thebibliography}

\end{document}